\documentclass[journal]{IEEEtran}

\usepackage{amsmath,amsfonts,bm}
\usepackage{array}
\usepackage{textcomp}
\usepackage{stfloats}
\usepackage{url}
\usepackage{verbatim}
\usepackage{graphicx}
\usepackage{cite}
\usepackage{amssymb}
\usepackage[ruled,linesnumbered, noend]{algorithm2e}
\usepackage{booktabs}
\usepackage{diagbox}

\ifCLASSOPTIONcompsoc
\usepackage[caption=false, font=footnotesize, labelfont=sf, textfont=sf,subrefformat=parens]{subfig}
\else
\usepackage[caption=false, font=footnotesize,,subrefformat=parens]{subfig}
\usepackage{makecell}
\usepackage{multirow}
\usepackage{xcolor}
\usepackage{tabularx}
\usepackage{bbding}
\usepackage{threeparttable}

\begin{document}

\title{Generative AI for Deep Reinforcement Learning: Framework, Analysis, and Use Cases}

\author{
        Geng~Sun,
        Wenwen~Xie,        Dusit~Niyato,~\IEEEmembership{Fellow,~IEEE},
        Fang~Mei, 
        Jiawen~Kang, 
        Hongyang~Du, 
        Shiwen~Mao,~\IEEEmembership{Fellow,~IEEE}
        \IEEEcompsocitemizethanks
        {
        \IEEEcompsocthanksitem This research is supported by the National Natural Science Foundation of China (62272194, 62471200), the Science and Technology Development Plan Project of Jilin Province (20220101101JC), the National Research Foundation, Singapore, and Infocomm Media Development Authority under its Future Communications Research \& Development Programme, Defence Science Organisation (DSO) National Laboratories under the AI Singapore Programme (FCP-NTU-RG-2022-010 and FCP-ASTAR-TG-2022-003), Singapore Ministry of Education (MOE) Tier 1 (RG87/22), and the NTU Centre for Computational Technologies in Finance (NTU-CCTF). (\textit{Corresponding author: Fang Mei.})
         \IEEEcompsocthanksitem Geng~Sun is with the College of Computer Science and Technology, Jilin University, Changchun 130012, China, and with Key Laboratory of Symbolic Computation and Knowledge Engineering of Ministry of Education, Jilin University, Changchun 130012, China, and also with the College of Computing and Data Science, Nanyang Technological University, Singapore 639798 (e-mail: sungeng@jlu.edu.cn).
        \IEEEcompsocthanksitem Wenwen~Xie and and Fang~Mei are with the College of Computer Science and Technology, Jilin University, Changchun 130012, China~(e-mail: xieww22@mails.jlu.edu.cn, meifang@jlu.edu.cn).
        \IEEEcompsocthanksitem Dusit~Niyato is with the College of Computing and Data Science, Nanyang Technological University, Singapore 639798 (e-mail: dniyato@ntu.edu.sg).
        \IEEEcompsocthanksitem Jiawen~Kang is with the School of Automation, Guangdong University of Technology, Guangzhou 510006, China (e-mail: kavinkang@gdut.edu.cn).
        \IEEEcompsocthanksitem Hongyang Du is with the Department of Electrical and Electronic Engineering, The University of Hong Kong, Hong Kong 999077, China~(email: duhy@hku.hk).
        \IEEEcompsocthanksitem Shiwen~Mao with the Department of Electrical and Computer Engineering, Auburn University, Auburn 36830, USA (e-mail: smao@ieee.org).
        }
 }

\maketitle

\begin{abstract}
As a form of artificial intelligence (AI) technology based on interactive learning, deep reinforcement learning (DRL) has been widely applied across various fields and has achieved remarkable accomplishments. However, DRL faces certain limitations, including low sample efficiency and poor generalization. Therefore, we present how to leverage generative AI (GAI) to address these issues above and enhance the performance of DRL algorithms in this paper. We first introduce several classic GAI and DRL algorithms and demonstrate the applications of GAI-enhanced DRL algorithms. Then, we discuss how to use GAI to improve DRL algorithms from the data and policy perspectives. Subsequently, we introduce a framework that demonstrates an actual and novel integration of GAI with DRL, \textit{i.e.}, GAI-enhanced DRL. Additionally, we provide a case study of the framework on UAV-assisted integrated near-field/far-field communication to validate the performance of the proposed framework. Moreover, we present several future directions. Finally, the related code is available at: https://xiewenwen22.github.io/GAI-enhanced-DRL.
\end{abstract}


%
\section{Introduction}
\par Deep Reinforcement Learning (DRL) is a transformative approach in the field of Artificial Intelligence (AI) that offers promising solutions to complex decision-making problems in various domains. DRL incorporates the principles of deep learning and reinforcement learning, where the introduction of neural networks has made DRL be capable of efficiently dealing with high-dimensional state and action spaces. Additionally, a DRL agent is able to learn directly from raw inputs without the need for artificial features and iteratively refine their strategies through trial-and-error interactions with an environment. This breakthrough capability has led to the extensive applications of DRL across diverse domains such as controlling robotic systems in dynamic environments.

\par Despite its remarkable achievements, traditional DRL algorithms suffer from certain limitations. One of the main challenges is the low sample efficiency for training deep neural networks in reinforcement learning environments. As the process of learning optimal policies often requires long and extensive interaction with an environment, data collection for DRL may be costly or impractical in the real world. Moreover, DRL models typically exhibit limited generalization capabilities, making it difficult to extend learned policies to unseen environments or tasks. 


\par Fortunately, Generative AI (GAI) can offer a promising solution to the challenges faced by DRL. Unlike discriminative AI, which aims to learn boundaries between different data categories, GAI can learn the distribution of data to capture its latent structure. Therefore, GAI is widely used in content analysis and creation domains. For example, ChatGPT\footnote{https://openai.com/index/chatgpt.} has revolutionized human-computer interaction by leveraging large language models (LLMs). Additionally, the recently introduced video generation model Sora\footnote{https://openai.com/index/sora.} has sparked a trend in the field of video generation. The widespread adoption of GAI can be attributed to several key advantages over other AI methods.
\begin{itemize}
    \item \textit{\textbf{High-quality Data Generation Ability}}: GAI can generate new credible data samples by learning the distribution of the data, which helps to solve the problem of data scarcity.
    \item \textit{\textbf{Feature Extraction Ability}}: GAI can extract useful features or representations from raw data by learning the distribution and latent structure of the data, which helps to improve the data analysis ability.
    \item \textit{\textbf{Generalization Ability}}: GAI can efficiently transfer the knowledge and representations acquired from one task to others, thereby accelerating the learning process and enhancing performance in new environments.
\end{itemize}

\begin{figure*}
    \centering
    \includegraphics[width=\linewidth]{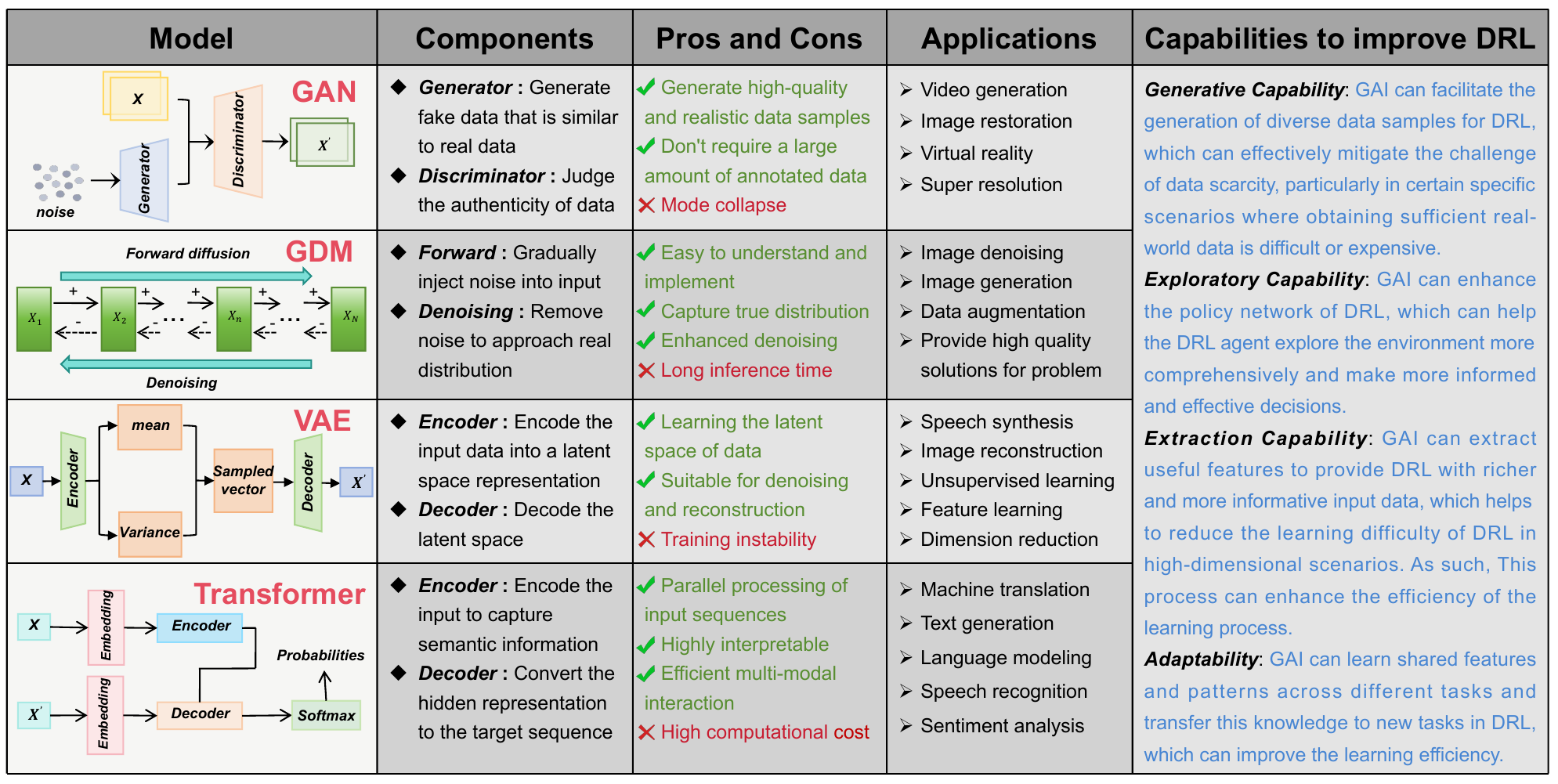}
    \caption{Summary of typical GAI models.}
    \label{fig:GAI}
\end{figure*}

\par Benefiting from the aforementioned advantages, GAI can be utilized to enhance DRL algorithms due to higher sample efficiency, stronger generalization capabilities, and robustness to environmental changes. For instance, a DRL agent may struggle in some specific environments, where the agent is not allowed to engage in extensive trial-and-error interactions within the environment. In this case, GAI can learn from limited training data to generate synthetic data for DRL training, which contributes to alleviating data scarcity and accelerating the learning process. Additionally, GAI can provide potential representations of data, which helps a DRL agent explore the environment more efficiently and generalize a policy better to unknown states. Therefore, the introduction of GAI brings new opportunities to DRL.

\par The combination of GAI and DRL is not straightforward and faces certain key issues. Specifically, GAI contains a variety of models that adept in solving different problems. Therefore, how to select an appropriate GAI model and integrate it with a specific DRL algorithm to solve certain problems is important and challenging. This is because the process involves the need for extensive studies and experimental validation of the compatibility between different models and algorithms to unleash full potential of GAI and DRL.

\par Motivated by this, we attempt to propose a comprehensive tutorial. \textit{To the best of our knowledge, this is the first work to comprehensively summarize how DRL algorithms can be enhanced using various GAI models.} The contributions can be summarized as follows:
\begin{itemize}
    
    \item We present the principles of important and widely used GAI models and DRL algorithms. Then, we briefly describe several applications of DRL enhanced with GAI.
    \item We explore different integration of GAI in DRL, focusing on how DRL can be enhanced from a data and strategy perspective. Moreover, we comprehensively summarize the advantages and disadvantages of the GAI-enhanced DRL algorithm.
    \item We propose a novel framework which deeply demonstrates the integration process of several common GAI models individually in the DRL algorithm. Moreover, we show the role of integrating multiple GAI models simultaneously in DRL to solve different challenges in DRL algorithms. Furthermore, we construct a case study to demonstrate the effectiveness of the proposed framework.
    
\end{itemize}


\section{Overview of GAI and DRL}
\par In this section, we introduce the basics of GAI and DRL.

\subsection{GAI Models}
\par The rapid development of GAI relies on several promising techniques, and we focus on the most commonly used Generative Adversarial Network (GAN), Generative Diffusion Model (GDM), Variational Autoencode (VAE), and Transformer. We summarize the technical details of GAI models and their comparison in Fig~\ref{fig:GAI}.

\begin{itemize}
    \item \textit{\textbf{GAN}}: GAN is a deep learning model composed of a generator and a discriminator. The former aims to generate samples similar to real data, while the latter attempts to distinguish between samples generated by the generator and real data. By adversarially training these two networks, the credibility of generated samples is gradually enhanced. The applications of GAN include image generation and style transfer.
    \item \textit{\textbf{GDM}}: The training of GDM consists of forward process and reverse process. Specifically, GDM destructs training data by gradually adding Gaussian noise in the forward process, then reverses this process to generate the desired data from the noise, which simulates the evolution of the data. The applications of GDM include image restoration and solving optimization problems.
    \item \textit{\textbf{VAE}}: VAE consists of an encoder and a decoder. The former maps input data to a probability distribution in the latent space, while the latter reconstructs samples from the latent space into the original data. Clearly, the higher the similarity between the original and reconstructed samples, the higher the model accuracy. VAE is widely applied in fields such as image generation and feature learning.
    \item \textit{\textbf{Transformer}}: Transformer is a deep learning model based on the self-attention mechanism, consisting of an encoder and a decoder. Specifically, the principle of the Transformer is to capture the dependencies between different positions in a sequence through self-attention mechanism, while enhancing representational ability of the model through multi-head attention mechanism. Transformer is widely used in the field of natural language processing, such as machine translation, language modeling.
\end{itemize}

\begin{table*}[]
  \centering
  \renewcommand\arraystretch{1.3}
  \caption{The Analysis of different DRL algorithms.}\label{tab:DRL}
  {
    \begin{tabularx}{0.77\textwidth} { 
       >{\hsize=.1\hsize\linewidth=\hsize\centering\arraybackslash}X 
      | >{\hsize=.12\hsize\linewidth=\hsize\centering\arraybackslash}X
      | >{\hsize=.40\hsize\linewidth=\hsize\centering\arraybackslash}X
      | >{\hsize=.332\hsize\linewidth=\hsize\centering\arraybackslash}X
      }
    \hline
    \multirow{2}{*}{\makecell[c]{\bf DRL \\ \bf Algorithm}} &
    \multicolumn{3}{c}{\bf Analysis} \\
    \cline{2-4}
      & \bf Policy & \bf Key Features & \bf Pros \& Cons \\
    \hline
    \bf DQN & Off-policy & \makecell*[c]{Estimating Q-function using DNN and \\ handle discrete action space} & \makecell*[l]{$\checkmark$ Simple to implement\\ $\checkmark$ Good generality \\ $\times$ Training instability} \\
    
    \cline{1-4}
    \bf DDPG & Off-policy & \makecell*[c]{Deterministic policy gradient method} & \makecell*[l]{$\checkmark$ Good stability \\ $\times$ Poor convergence} \\
    
    \cline{1-4}
    \bf TD3 & Off-policy & \makecell*[c]{Dual-Q network and delayed policy \\ network updates} & \makecell*[l]{$\checkmark$ Improved training stability \\ $\checkmark$ Reduced overfitting \\ $\times$ Hyperparameter sensitivity} \\
    
    \cline{1-4}
    \bf PPO & On-policy & \makecell*[c]{Limiting the magnitude of policy \\ network updates} & \makecell*[l]{$\checkmark$ High stability \\ $\checkmark$ High training efficiency \\ $\times$ Low sample efficiency} \\
    
    \cline{1-4}
    \bf SAC & Off-policy & \makecell*[c]{Entropy regularization terms and \\ dual-policy networks} & \makecell*[l]{$\checkmark$ Explore action space more fully \\ $\checkmark$ Insensitive to parameters \\ $\times$ High training complexity} \\
    \hline
    \end{tabularx}
    }
\end{table*}

\subsection{DRL Algorithms}
\par Reinforcement Learning (RL) is a machine learning method that aims to enable an agent to maximize cumulative rewards through trial-and-error learning during their interaction with the environment. Specifically, the agent learns to make optimal decisions in a given environment based on the state and feedback rewards of the environment. Note that traditional RL algorithms use tables to estimate the Q-function that helps the agent to improve policy. However, it is clearly not suitable for high-dimensional and large state and action spaces. As such, DRL combining deep neural networks (DNN) and RL has been proposed, where DNN is used as an approximator of the Q-function. We summarize several representative DRL algorithms, namely, deep Q-network (DQN), deep deterministic policy gradient (DDPG), twin delayed deep deterministic policy gradient (TD3), proximal policy optimization (PPO), and soft actor-critic (SAC). The details of the DRL algorithms above are shown in Table~\ref{tab:DRL}.
    
    
    
    

\par Considering the ability of DRL to handle dynamic problems and find optimal solutions in the absence of information, it has achieved remarkable results in data communications and networking. However, DRL still has some limitations that may be mitigated by GAI:
\begin{itemize}
    \item \textit{Sample Inefficiency}: DRL typically requires an extensive interaction with the environment to learn effective policy. In this case, GAI (\textit{e.g.}, GAN) can help generate synthetic data to augment training dataset, which improves sample efficiency~\cite{GAN+PPO+DataGeneration}.
    
    \item \textit{Challenges in Complex Environment}: A DRL agent may fall into local optima or struggle to explore large and complex environments efficiently. In this instance, GAI (\textit{e.g.}, VAE) enables the agent to analyze the underlying data distribution and perform feature extraction, which helps to accurately model the environment and improve the stability of training process~\cite{VAE+DDPG+lowDimensional}. 
    
    \item \textit{Instability and Low Learning Efficiency}: DRL may suffer from instability and balance of exploration-exploitation trade-off. In this situation, GAI (\textit{e.g.}, GDM) can be used to improve the policy network of DRL, which helps to enhance the stability of exploration~\cite{Diffusion+SAC}.
\end{itemize}



\subsection{Applications of GAI-enhanced DRL}
\par Thanks to the excellent generative and analytical capabilities of GAI, it has been widely applied to enhance DRL in various fields. We have summarized typical applications.
\begin{itemize}
    \item \textit{Signal Processing}: Training an agent to solve control tasks directly from multimodal signals has been shown to be difficult. In~\cite{Transformer+DDPG+Multimodal}, a framework combining Transformer and DRL is proposed, where Transformer is used for feature extraction of multimodal data, which can better integrate important information carried by different signals to assist the agent in making more accurate decisions.

    \item \textit{Wireless Communications}: GAI can be utilized to ameliorate the problem of high training cost and low sample efficiency against DRL in wireless environments. In~\cite{GAN+TD3+EstimateEnvironment}, the authors proposed a GAN-based auxiliary training mechanism that reduces the overhead of interacting with the real-world by generating environment states for offline training.
    
    \item \textit{Edge-based AIGC Service}: GAI combined with DRL is utilized to provide edge-based AIGC services. For example, the authors in~\cite{Diffusion+SAC} proposed a diffusion model based approach to generate optimal AIGC service provider selection decisions and combined it with SAC algorithm to improve the efficiency and effectiveness of the algorithm.
    

    \item \textit{Sensor Networks}: GAI can be utilized to improve the accuracy of DRL decision making for cognitive internet of things network. In~\cite{Lin2024}, GAI is employed to enhance the value network of DRL by improving the stability and accuracy of action value estimation through adversarial training, thereby enhancing the decision-making performance of DRL.
    
\end{itemize}

\par \textit{\textbf{Lesson Learned}}: From the above analysis of the algorithms and applications, we observe that GAI models can effectively address the limitations of DRL. Specifically, the data generation capability of GAN contributes to tackling the data scarcity issue in certain scenarios. VAE can be employed for extracting features of high-dimensional data. Moreover, the exploration capability of GDM can be utilized to enhance the policy network of DRL, thereby improving the quality and stability of decision-making. The proficiency of Transformers in handling multimodal and variable-length data expands the scope of applications for DRL.

\begin{figure*}
    \centering
    \includegraphics[width=\linewidth]{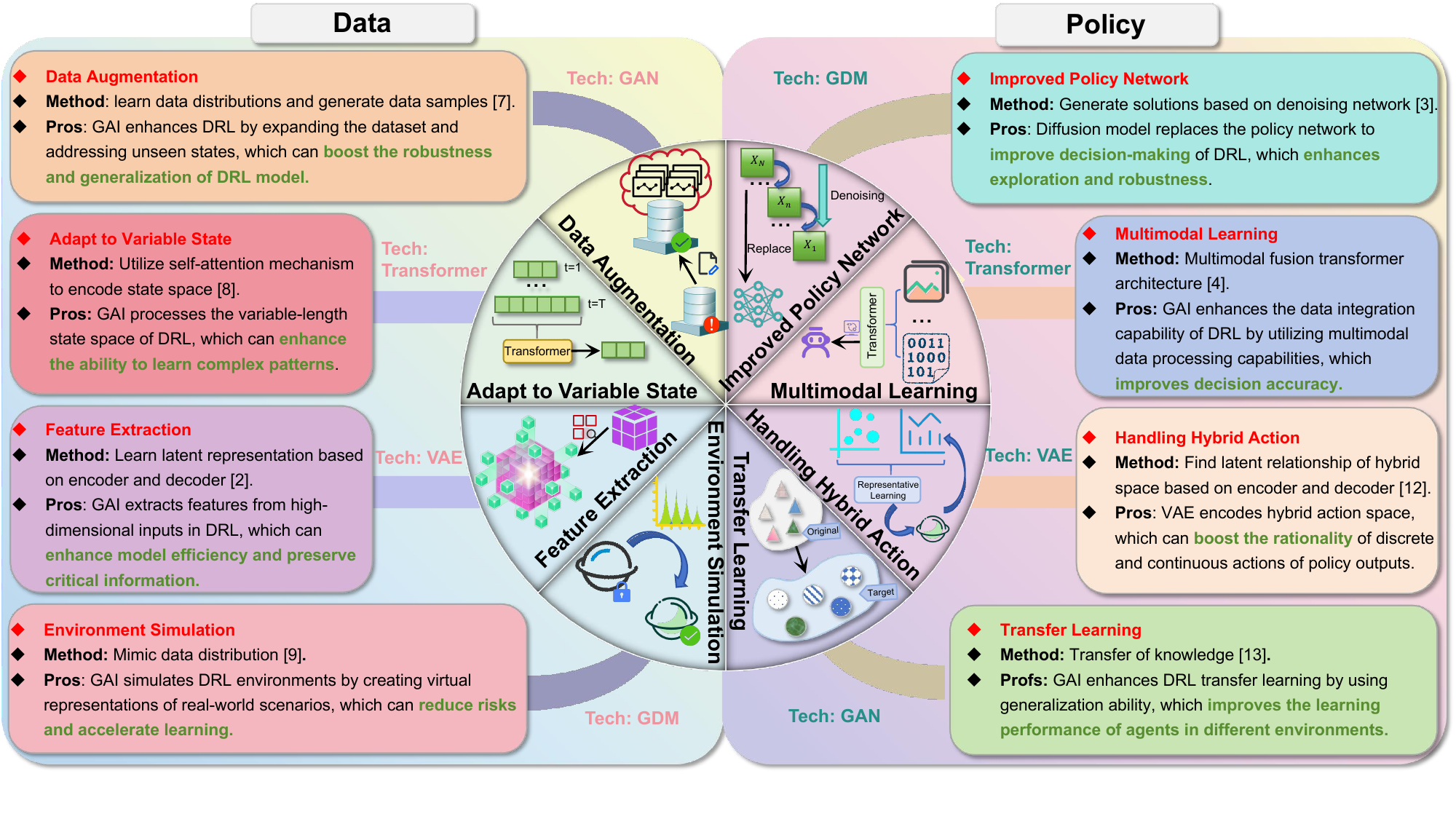}
    \caption{Summary of data and policy performance improved by GAI in DRL. }
    \label{fig:GAI_DRL}
\end{figure*}


\section{GAI-enhanced DRL}
\label{GAI-enabled DRL}

\par In this section, we introduce how GAI can improve DRL from the data and policy perspective. The summary of improvements are shown in Fig.~\ref{fig:GAI_DRL}.

\subsection{From Data Perspective}
\par Based on aforementioned unique features, GAI can effectively address the data-related challenges to improve the convergence speed and solution accuracy of DRL.

\subsubsection{Data Augmentation}
\par In most DRL training processes, agents sample experiences stored in an experience replay buffer  randomly or in a specific way (\textit{e.g.}, prioritized experience replay) and utilize these experiences to update neural networks for better performance. Simultaneously, it is challenging for agents to output high-quality actions based on unobserved states or with limited data in such training approach. Adopting GAI to augment data can be a solution. For example, in~\cite{GAN+DDPG+ExperienceGeneration}, the authors proposed a GAN-powered multi agent DDPG (MADDPG) framework for joint optimizing terminal-cooperative caching and offloading for virtual reality tasks. Specifically, given high costs and privacy risk issues associated with collecting data from multiple users in a decentralized network, GANs are used to learn the distribution of real data and generate appropriate data for each agent based on the state. Then, each agent inputs the corresponding state samples generated by GAN into the actor-critic network and stores the resulting experiences in the experience buffer for sharing, which augments the agent with limited samples of real virtual reality data. Numerical results indicate a 5.32\% enhancement in energy efficiency for GAN-MADDPG compared with MADDPG trained on a real environment, with similar performance in terms of task completion latency.

\subsubsection{Adaptation to variable state} 
\par DRL with a standard deep neural network typically deals with fixed-length state spaces, which makes it challenging to handle problems with variable-length state spaces. However, in certain scenarios such as edge computing and task offloading, dynamic state spaces are relatively common, which limits the applicability of DRL in such contexts. Note that GAI is excellent in dealing with variable-length inputs, especially Transformer. A Transformer-based multi-agent DRL framework for scalable multi-unmanned aerial vehicle (UAV) area coverage was proposed~\cite{Transformer+DDPG+VariableState}. In such a scenario, a deep neural network based on a fixed-dimension multilayer perceptron (MLP) is no longer applicable. This is because when the number of UAVs varies, some dimensions of the state space are blank, which may lead to neural network errors. In this case, Transformer is adopted to encode the state space and assign different weights to deal with variable input dimensions through a self-attention mechanism, which can mitigate the effect of blank states on decision making. Meanwhile, there is a large amount of heterogeneous information in multi-UAV scenarios, where the importance of this information varies. The self-attention mechanism can dynamically learn the importance of different positions in the input sequence, which empowers the agent to extract most relevant information. Experimental results show that Transformer-based DDPG achieves the best network performance. Compared with the expert coverage-first algorithm, the Transformer-based DDPG algorithm improves the average coverage score by 39.7\% and the fairness index by 45.3\%.

\subsubsection{Feature Extraction}
\par When dealing with high-dimensional data, DRL requires a significant amount of resources and time. Therefore, it is important for DRL to reduce the dimension of data without losing important information. Fortunately, GAI can analyze the latent relationships in data and perform feature extraction. Therefore, GAI can be used to process input data of DRL to alleviate the limitations of DRL when facing high-dimensional data. For instance, in~\cite{VAE+DDPG+lowDimensional}, the authors proposed an environment perception framework for autonomous driving, where VAE is combined with SAC to focus on uninterrupted and reasonably safe autonomous driving. Specifically, VAE is used to process input driving environment images for feature extraction, which improves sample efficiency and facilitates learning processes with fewer samples but higher robustness. Simulation results on the DonKey simulator demonstrate the VAE-SAC method enables an autonomous vehicle to remain on track for the maximum time in a given timeframe. Note that although both VAE and Transformer can be used to handle the state space of an agent, their focuses are different. VAE~\cite{VAE+DDPG+lowDimensional} primarily aimes at improving sample efficiency by extracting features to reduce the dimensionality of inputs. Conversely, Transformer~\cite{Transformer+DDPG+VariableState} is not used explicitly for dimensionality reduction, and its main purpose is to handle variable-length state spaces through focusing on important parts of the input sequence.

\subsubsection{Environment Simulation}
\par Agent-environment interaction is a crucial aspect of DRL training processes. However, in certain scenarios, an agent are not allowed to perform trial-and-error training in real environments. Therefore, considering that GAI is able to learn and model data distribution, it can be harnessed to simulate the environment of DRL. For example, the authors in~\cite{GAN+SAC+EstimateEnvironment} proposed a primary-user-friendly dynamic spectrum anti-jamming framework. Note that it is inevitable for a second user (SU) to interfere the primary user (PU) even DRL is used by the SU to optimize spectrum access, which is undesirable in overlay cognitive radio network. Therefore, a GAN-based virtual environment (VE) is utilized to accurately simulate the spectrum environment. Subsequently, a channel decision network (CDN) based on DRL learns an optimal spectrum access strategy in VE through offline training. Experimental results show that the proposed framework converges much faster than the scheme that trains CDNs in the spectrum environment from scratch.


\par \textit{\textbf{Lesson Learned}}: From the studies above, we summarize the uses of GAI to improve DRL from the data perspective as follows.
\begin{itemize}
    \item \textit{Data Generation}: Given the powerful ability to mimic data distributions and generate credible data samples, GAI can be used to expand training datasets and anticipate unknown situations~\cite{GAN+DDPG+ExperienceGeneration,GAN+SAC+EstimateEnvironment}.
    \item Data Processing: GAN can be used to process dynamic data and high-dimensional data with unique analyzing and learning capabilities~\cite{Transformer+DDPG+VariableState,VAE+DDPG+lowDimensional}.
\end{itemize}

\subsection{From Policy Perspective}
\par By fully leveraging its ability to extract latent patterns and adapt to various data types, GAI can enhance the decision-making capability of DRL, enabling it to tackle complex tasks more efficiently and flexibly.

\subsubsection{Improved Policy Network}
\par The structure of a DRL policy network can affect quality of the output action. Currently, GAI has been used to enhance DRL policy network, especially in diffusion model. In~\cite{Diffusion+SAC}, the authors proposed a diffusion model-based SAC algorithm to generate an optimal AIGC service provider (ASP) selection decision for better and broader AIGC services in wireless networks. Specifically, the diffusion model is utilized to replace the policy network of SAC to generate action distributions based on observed states. It is worth noting that optimization problems in wireless networks usually are difficult to achieve optimal solutions. Thus, only the reverse process of the diffusion model is employed to construct the policy network for SAC. Extensive experimental results demonstrate the effectiveness of the proposed diffusion model-based algorithm, which outperforms seven representative DRL algorithms (\textit{e.g.}, PPO) in both the ASP selection problem and various standard control tasks.

\subsubsection{Multimodal Learning}
\par DRL is typically limited to handling a single type of data, which restricts its applicability. The GAI's ability of processing multimodal data can be employed to enhance DRL. In~\cite{Transformer+DDPG+Multimodal}, the authors proposed a Transformer-based DDPG to solve autonomous vehicle decision problem in complex scenarios. Multimodal Transformer is utilized to handle different types of data, \textit{i.e.}, LiDAR point cloud and images. Then, DDPG is employed to complete the subsequent autonomous driving decision-making task based on the extracted features. The experimental results indicate that the DDPG based on multimodal Transformer achieves higher rewards than DDPG that uses only images, which suggests that multimodal Transformer plays a significant role in extracting key multi-modal features.

\begin{table*}[]
  \centering
  \renewcommand\arraystretch{1.3}
  \caption{The Use of GAI for Various DRL Algorithms}\label{tab:dummy-1}
  {
    \begin{tabularx}{\textwidth} { 
       >{\hsize=.07\hsize\linewidth=\hsize\centering\arraybackslash}X 
      | >{\hsize=.17\hsize\linewidth=\hsize\centering\arraybackslash}X 
      | >{\hsize=.17\hsize\linewidth=\hsize\centering\arraybackslash}X
      | >{\hsize=.17\hsize\linewidth=\hsize\centering\arraybackslash}X
      | >{\hsize=.17\hsize\linewidth=\hsize\centering\arraybackslash}X
      | >{\hsize=.25\hsize\linewidth=\hsize\centering\arraybackslash}X
       }
    \hline
    \multirow{2}{*}{\diagbox[width=5em]{\bf DRL}{\bf GAI}} &
    \multicolumn{4}{c|}{\bf GAI Models} &
    \multirow{2}{*}{\bf Anaylsis} \\
    \cline{2-5}
      & \bf Transformer & \bf GDM & \bf GAN & \bf VAE & \\
    \hline
    \bf DQN &  \makecell*[l]{$\bullet$ Introduce the \\ transformer structure \\ into policy neural \\ networks.} & \makecell*[l]{$\bullet$ Model in-support \\ trajectory sequences. } & \makecell*[l]{$\bullet$ Apply the GAN \\structure to actor and \\critic networks~\cite{GAN+DQN+ImprovePolicy}.} & \makecell*[l]{$\bullet$ Capture the low \\ dimensional latent states \\ for facilitating DQN \\ agent learning.} & \multirow{5}{*}{\makecell*[l]{\textbf{Potential Benefits}:\\ \\ 1) \textbf{Expressiveness}: GAI is able to \\ learn complex sample distributions \\ well, which enables it to be \\ integrated into the policy network \\ of DRL for decision-making. \\ \\ 2) \textbf{Sample Quality}: GAI has strong \\ imitation and generation abilities, \\ allowing it to generate highly \\ credible new samples based on \\ existing samples to expand the \\ training dataset of DRL. \\ \\ 3) \textbf{Flexibility}: The ability of GAI \\ to model diverse behaviors is \\ particularly useful in DRL, where \\ the agent needs to adapt to a \\ variety of situations and tasks. }}\\
    
    \cline{1-5}
    \bf DDPG & \makecell*[l]{$\bullet$ Process multimodal \\ data~\cite{Transformer+DDPG+Multimodal}. \\ $\bullet$ Handle variable \\ state space~\cite{Transformer+DDPG+VariableState}.} & \makecell*[l]{$\bullet$ Introduce the GDM \\ structure into policy \\ networks~\cite{Diffusion+DDPG}.} & \makecell*[l]{$\bullet$ Generate complete \\ dataset including not \\ appeared states~\cite{GAN+DDPG+ExperienceGeneration}.} & \makecell*[l]{$\bullet$ Provide the extracted \\ feature state as input \\ for policy network~\cite{VAE+DDPG+lowDimensional}.} & \\
    
    \cline{1-5}
    \bf TD3 & \makecell*[l]{$\bullet$ Introduce transformer \\ module to the actor \\network.} & \makecell*[l]{$\bullet$ Approximate actions \\ by noise networks to \\ improve the speed of \\ training.} & \makecell*[l]{$\bullet$ Generate environment \\ states for offline training \\ of TD3 agent~\cite{GAN+TD3+EstimateEnvironment}.} & \makecell*[l]{$\bullet$ Provide the latent \\ representation of \\ continuous-discrete \\ hybrid action space~\cite{VAE+TD3+HybridAction}.} & \\
    
    \cline{1-5}
    \bf PPO & \makecell*[l]{$\bullet$ Employ transformer \\ structure as the police \\network.} & \makecell*[l]{$\bullet$ Generate samples.} & \makecell*[l]{$\bullet$ Generate new training \\ data based on historical \\ data~\cite{GAN+PPO+DataGeneration}.} & \makecell*[l]{$\bullet$ Encode input video \\ to reduce the state \\ dimension.} & \\
    
    \cline{1-5}
    \bf SAC & \makecell*[l]{$\bullet$ Extract latent  \\features. \\ $\bullet$ Encode historical \\ information and hybrid \\ space.} & \makecell*[l]{$\bullet$ Apply GDM structure \\ to policy network~\cite{Diffusion+SAC}. \\ $\bullet$ Simulate expert \\ behavior to generate \\ optimal policy.} & \makecell*[l]{$\bullet$ Provide a simulated \\ environment for SAC \\ agent~\cite{GAN+SAC+EstimateEnvironment}.} & \makecell*[l]{$\bullet$ Map high-dimensional \\ states into the latent \\ vectors. \\ $\bullet$ Encode tasks and \\ infer tasks.} & \\
    \hline
    \end{tabularx}
    }
\end{table*}

\subsubsection{Handling hybrid action}
\par DRL is typically used to deal with hybrid action spaces containing both discrete and continuous actions, especially for optimization problems involving scheduling. Current common practices either approximate the hybrid space by discretization or relax it into a continuous set. 
However, continuous actions and discrete actions typically interact with each other in general, while the aforementioned discretization or relaxation methods ignore this intrinsic relationship and leads to information loss. In this case, GAI constructs latent space representations and analyze the dependencies of hybrid action space. The authors in~\cite{VAE+TD3+HybridAction} proposed a novel VAE-based DRL framework called HyAR. The main idea of the proposed framework is to construct a unified and decodable representation space for the original discrete-continuous hybrid action, in which the agent learns a latent strategy. Then, the selected latent action is decoded back into the original hybrid action space to interact with the environment. The results show that HyAR has advantages over the baseline (\textit{e.g.}, PADDPG), especially in high-dimensional action space.

\subsubsection{Transfer Learning}
\par One of the major dilemmas of DRL is that if the structure of network states change dramatically, we have to reformulate the problem and retrain the model from scratch. Fortunately, GAI can be used for knowledge transfer, which can alleviate the limitations. For example, in~\cite{Wang2023}, the authors proposed a diffusion model-based transfer learning method to solve the limited data problem in image generation tasks. The proposed method trains a diffusion model with an adversarial noise selection and a similarity-guided strategy, which improves the efficiency of the transfer learning process. Extensive experiments in the context of few-shot image generation tasks demonstrate that the proposed transfer learning method is not only efficient but also excels in terms of image quality and diversity. Inspired by this work, we can adopt GAI model (\textit{e.g.}, diffusion model and GAN) to design transfer learning for DRL.


\par \textit{\textbf{Lesson Learned}}: We find that the GAI approaches to improve policy performance can be summarized into two aspects:
\begin{itemize}
    \item Independent Strategy: GAI can be regarded as an independent strategy to preprocess the inputs of DRL, which can enhance the training efficiency of DRL~\cite{Transformer+DDPG+Multimodal}.
    \item Integrated Strategy: GAI can be integrated into DRL architecture to enhance the solution capability of DRL~\cite{Diffusion+SAC}.
\end{itemize}

\begin{figure*}
    \centering
    \includegraphics[width=\linewidth]{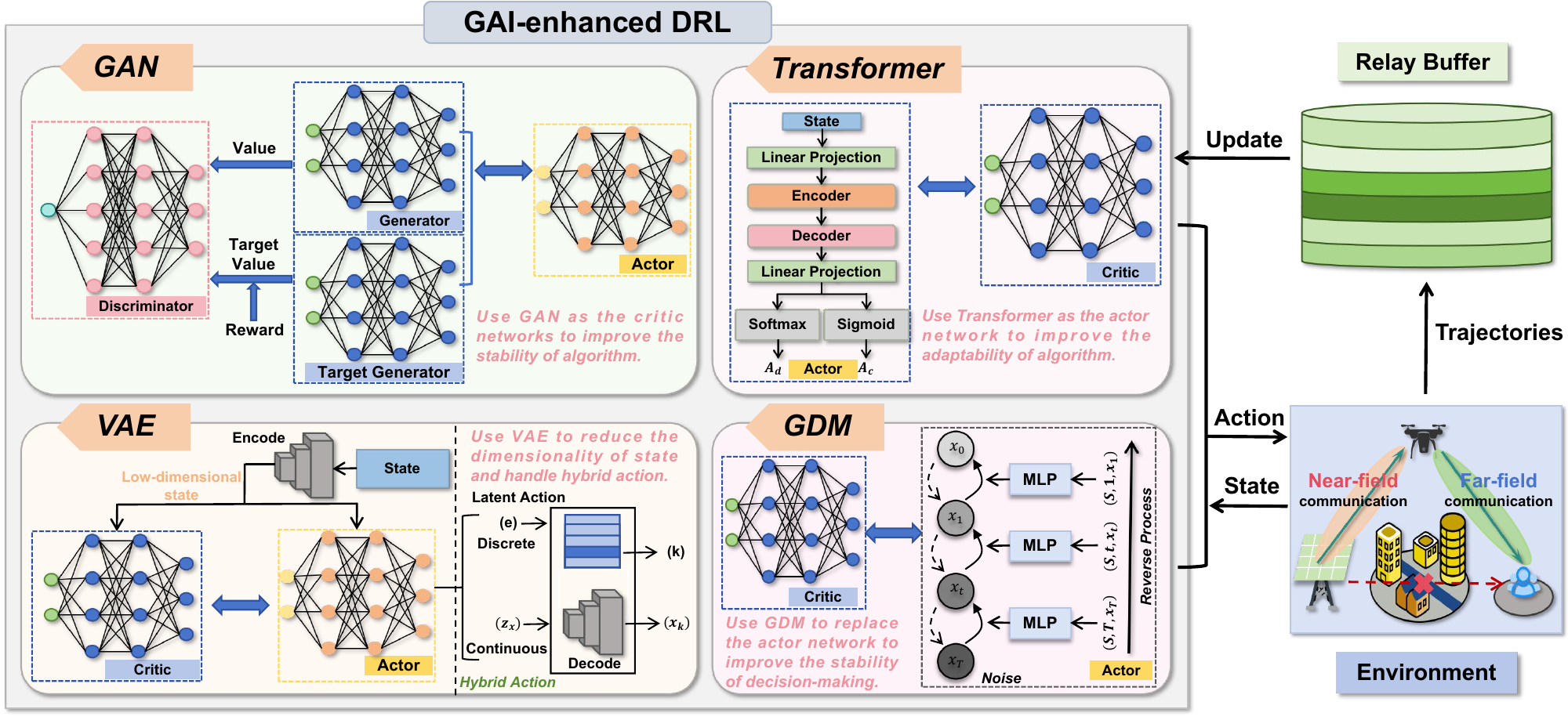}
    \caption{Proposed Framework of GAI-enhanced DRL on UAV-assisted integrated near-field/far-field communication.}
    \label{fig:Framework}
\end{figure*}


\subsection{Discussion}

\par From the applications shown in Table~\ref{tab:dummy-1}, we can find that integrating GAI into DRL can significantly improve the performance of DRL, which can be summarized as follows:
\begin{itemize}
    \item \textit{Improved Data Diversity}: The generative and learning capability of GAI can increase the diversity of training data for DRL to alleviate the data scarcity problem in certain particular scenarios, which can improve the speed of DRL training. In addition, the multimodal data processing capability of GAI greatly broadens the application scope of DRL.

    
    \item \textit{Enhanced Analysis Ability}: Some GAI models, such as VAEs, are particularly effective for high-dimensional data, as it can eliminate the influence of redundant information and enhance the speed of DRL training.

    
    \item \textit{Improved Decision Ability}: The reverse process of the diffusion model is well-suited to the policy networks of DRL, where multi-round iterations and denoising processes help agents explore better actions. Additionally, GAI has inspired the implementation of transfer learning in DRL to facilitate interactions across similar domains.

\end{itemize}
\par Although incorporating GAI into DRL addresses some issues faced by DRL, it also introduces some disadvantages as follows:
\begin{itemize}
    \item \textit{High Computation Complexity}: Integrating GAI models into DRL algorithms may introduce additional computational costs, such as tuning the network architecture of GAI models and a large number of hyperparameters, which can increase computational overhead during training. Moreover, the training process of GAI models requires iterative optimization, which leads to an increase in the overall training time of DRL.
    \item \textit{High Resource Demand}: GAI models typically require substantial memory to store model weights, parameters, and intermediate results, which is particularly pronounced in data-intensive applications. Moreover, GAI models often have complex structures and require high-performance computing resources to support their training when performing computationally intensive tasks.
\end{itemize}

\par Therefore, GAI-enhanced DRL is suitable for solving the issues of low sample efficiency and insufficient exploration by generating diverse training data and high-quality actions with sufficient computational resources.

\section{Case Study: GAI-enhanced DRL in Near-field Communication}

\par In this section, we first propose a novel framework for GAI-enhanced DRL. We then introduce a case study to demonstrate effectiveness of the proposed framework.


\subsection{Proposed Framework}
\par As shown in Fig.~\ref{fig:Framework}, we investigate an integration of various GAI models into DRL algorithms to enhance the performance in different ways. 
\subsubsection{GAN-enhanced DRL} We enhance the critic network of DRL by using GAN. Specifically, the generator network outputs estimated action values, while the target generator network obtains the target action values. The discriminator network attempts to minimize the distance between the estimated action values and the target action values calculated by the Bellman operator. Introducing adversarial learning mechanism inherent in GAN to approximate the action-value distribution avoids the negative impact of randomness on rewards, which enhances the stability of DRL.
\subsubsection{VAE-enhanced DRL} We use VAE to reduce the dimensionality of the high-dimensional state space in DRL. Specifically, we input the high-dimensional state data into the encoder of the VAE to extract a lower-dimensional latent variable. Subsequently, the decoder attempts to reconstruct the original state data from the latent variable. By minimizing the discrepancy between the reconstructed data and the original data, the VAE is able to maintain the quality of the lower-dimensional latent variable while performing dimensionality reduction. Additionally, VAE can construct a latent representation space for continuous parameters conditioned on state and embedding of discrete actions to handle hybrid actions.

\subsubsection{Transformer-enhanced DRL} We enhance the actor network of DRL by using Transformer. Specifically, we replace the MLP with a network based on the attention mechanism of Transformer to analyze the current state in the environment. This enables Transformer-enhanced DRL to more effectively capture long-range dependencies within sequences, making it well-suited for decision-making tasks that require long-term memory or complex historical information.

\subsubsection{GDM-enhanced DRL} We improve the policy network of DRL by employing the reverse process of GDM. Specifically, we consider the policy network as a denoiser, progressively adding denoising noise to the initial Gaussian noise to recover or discover the optimal actions. Benefiting from the generative ability and expressiveness of GDM, it can generate diverse and high-quality actions, which makes it suitable for improving DRL algorithms with limited exploration capabilities.


\begin{figure*}
    \centering
    \includegraphics[width=\linewidth]{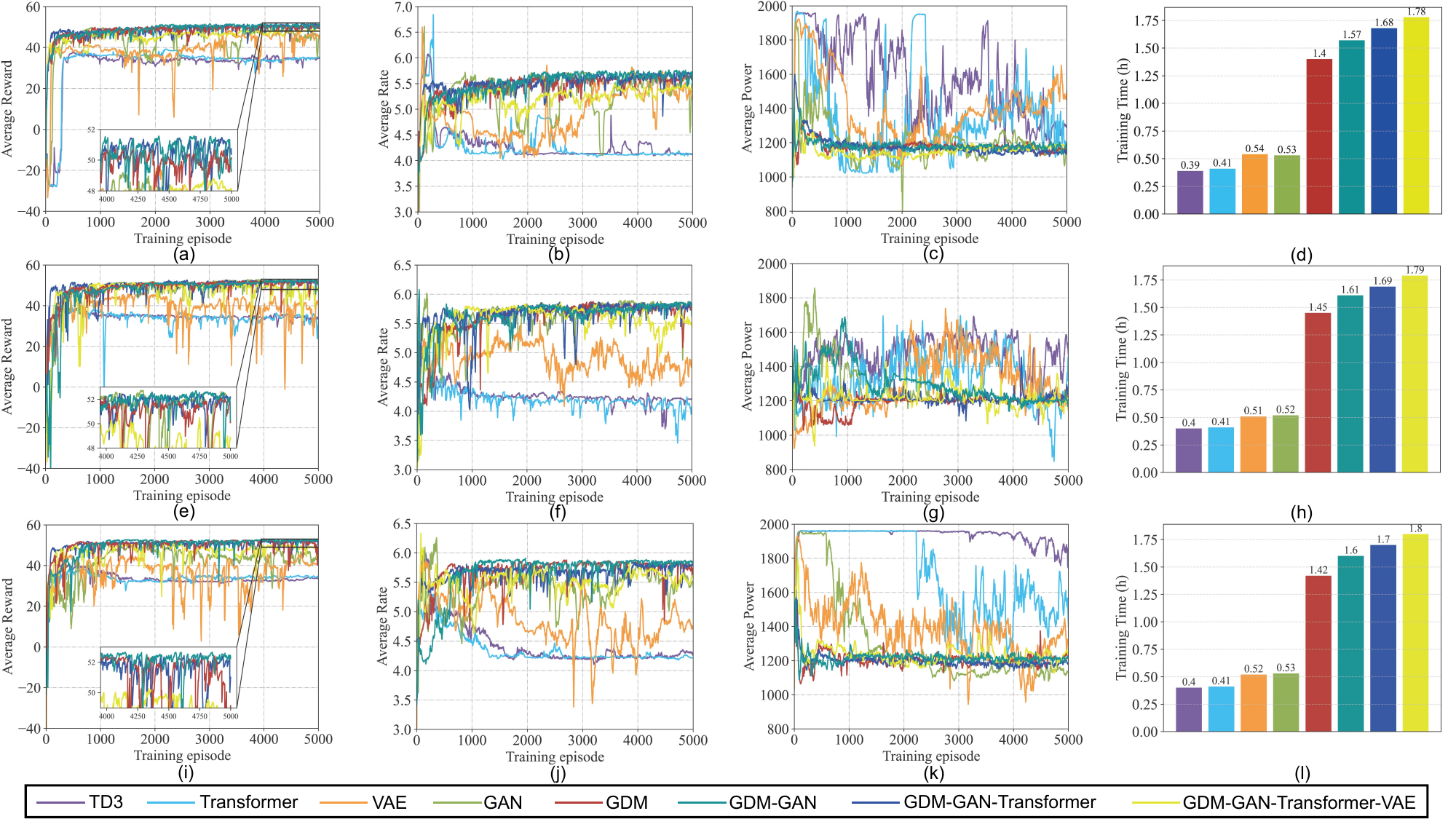}
    \caption{The convergence curves of rewards, rates, and power consumption, as well as the training time, of the case study under three different action spaces. Specifically, the results in continuous action space are shown in (a), (b), (c) and (d), respectively. The results in discrete action space are shown in (e), (f), (g) and (h), respectively. The results in hybrid action space are shown in (i), (j), (k) and (l), respectively.
    }
    \label{fig:performance of case study}
\end{figure*}


\subsection{Scenario Description}

\par Large-scale antenna array has demonstrated an important enhancement in the spectrum efficiency of wireless systems. It is noteworthy that with the significant increase in carrier frequency and the number of antennas, the well-known Rayleigh distance expands to tens or even hundreds of meters. In this context, near-field communication becomes more prominent~\cite{Wu2024}. Unlike traditional far-field communication where the electromagnetic field is simply modeled by plane waves, the electromagnetic field in near-field communication needs to be accurately modeled by spherical waves. Therefore, we can exploit this unique channel characteristic to achieve flexible beamforming, concentrating the beam energy at specific spatial locations rather than focusing it in a specific direction as in traditional far-field communication. The characteristic above provides near-field communication with distinct advantages in improving signal quality, especially in densely populated urban areas.



\par In the case study, we considered a UAV-assisted ground communication, where the UAV serves as a relay to transmit data from a base station (BS) equipped with a large-scale antenna array to the distant ground user. Here, the UAV is deployed and located in the near-field region of BS while the ground users are in the far-field region of the UAV. In this case, we aim to maximize the transmission rate and minimize the power consumption by jointly optimizing the trajectory of the UAV and the power allocation of BS and UAV.



\subsection{Result Analysis}
\subsubsection{Analysis of Different Types of Action Spaces}
\par We explore the performance of GAI-enhanced DRL in continuous action space (\textit{i.e.}, continuous UAV trajectory design and power allocation), discrete action space (\textit{i.e.}, way-point discrete UAV trajectory design and power allocation), and hybrid action spaces (\textit{i.e.}, discrete UAV trajectory design and continuous power allocation).
\par \textbf{Continuous Action Space}: Figs.~\ref{fig:performance of case study}(a), (b) and (c) show the convergence curves of reward, transmission rate and power consumption in continuous action space, respectively. As an improved version of DDPG, TD3 significantly enhances policy stability and performance through the use of twin Q-networks and delayed policy updates. Therefore, in our case study, we select TD3 as the foundational DRL algorithm. We can observe that the GAN-enhanced TD3 algorithm shows significant improvements over the original TD3 algorithm, and this improvement is due to the novel adversarial training of the critic and target critic networks, which enhances the process of state-action value estimation. However, it is unable to effectively learn the constraints on power within the continuous solution space, resulting in penalties that cause fluctuations in the reward. Moreover, we observe that VAE improves the performance of the TD3 algorithm, which is attributed to the capability of VAE to reduce the dimensionality of the high-dimensional state space to extract critical information. However, the process of transforming the high-dimensional state space into a low-dimensional latent space inevitably results in loss of information and leads to instability.. Furthermore, we find that the Transformer-enhanced TD3 algorithm barely improves the performance of the original TD3 algorithm. This may be because Transformer is better suited for handling long-range problems, which may not align well with the current optimization problem. In addition, we observe that GDM-enhanced-TD3 achieves the best reward values and transmission rates compared to other algorithms. This is because GDM can accurately capture the underlying data distribution, providing a more effective representation of the environment. Moreover, the unique structure of GDM, which involves a diffusion process, offers a more stable and efficient learning process.

\par \textbf{Discrete Action Space}: The convergence curves of reward, transmission rate and power consumption in the discrete action space are shown in Figs.~\ref{fig:performance of case study}(e), (f) and (g), respectively. Unlike its performance in continuous action space, GAN-enhanced GAI demonstrates superior convergence speed and quality in all of metrics above. This is because the discrete action space we designed inherently satisfies the power constraints of the optimization problem, eliminating the need for GAN to learn these constraints. However, this also indirectly highlights the weakness of GAN in learning constraint conditions directly related to optimization variables. Moreover, the convergence curves of VAE-enhanced GAI in discrete space are relatively unstable. This could be due to the fact that in discrete action spaces, the decision boundaries are usually more complex and require precise state information to make correct action choices. However, the latent space obtained after dimensionality reduction by the VAE model may not provide sufficiently accurate information to correctly distinguish different discrete actions, leading to decision biases. Furthermore, we observe from the figure that the GDM-based algorithm still achieves the excellent performance, which indicates that it can handle discrete actions well.

\par \textbf{Hybrid Action Space}: Figs.~\ref{fig:performance of case study}(i), (j) and (k) show the convergence curves of reward, transmission rate and power consumption in hybrid action space, respectively. We can observe that the GDM-enhanced TD3 is optimal in terms of the convergence speed and stability on the hybrid action space, and it converges to the best values with regard to reward, transmission rate, and power consumption. In the hybrid action space, the discrete and continuous actions are not independent, and different discrete actions correspond to different optimal continuous actions~\cite{VAE+TD3+HybridAction}. Therefore, to make the best hybrid action decisions, it is essential to explore the latent relationships between discrete and continuous actions. Evidently, the strong capability of GDM in learning the underlying data relationships enhances the accuracy and stability of TD3 decisions, enabling GDM-enhanced TD3 to achieve outstanding performance in the hybrid action space. 

\subsubsection{Analysis of Multiple GAI Models Combination}
\par We further combine multiple GAI models in TD3 to explore their performance on the abovementioned three action spaces. Specifically, we propose the GDM-GAN-enhanced TD3 algorithm, which introduces GAN into the GDM-enhanced TD3 algorithm to improve the critic network. Based on this, we design the GDM-GAN-Transformer-enhanced TD3 algorithm, which improves the denoising network of GDM by using Transformer. Finally, the GDM-GAN-Transformer-VAE-enhanced TD3 algorithm is proposed, which introduces VAE into the GDM-GAN-Transformer-enhanced TD3 to reduce the dimensionality of the state space. The results are shown in Fig.~\ref{fig:performance of case study} and corresponding results analysis are as follows:

\begin{itemize}
    \item \textit{Results of GDM-GAN-enhanced TD3 algorithm}: As can be seen, the GDM-GAN-enhanced TD3 algorithm slightly outperforms the GDM-enhanced TD3 algorithm. This is because the GAN further enhances the quality of the estimation of state-action values by the critic network, which is beneficial for the agent to explore more effective actions.
    
    \item \textit{Results of GDM-GAN-Transformer-enhanced TD3 algorithm}: Introducing the Transformer does not lead to significant changes in the performance of the GDM-GAN-enhanced TD3 algorithm. This is because Transformer is more suitable for addressing long-range optimization problems and it does not provide substantial performance gains in our considered case study.
    
    \item \textit{Results of GDM-GAN-Transformer-VAE-enabled TD3 algorithm}: Adopting VAE results in the performance decline of the GDM-GAN-Transformer-enhanced TD3 algorithm. This decline results from the information loss caused by the dimensionality reduction of state data using VAE, which negatively impacts the GAN-based critic network that relies on detailed state information, leading to an overall performance decrease.
    
    \item \textit{Training Time}: The training time is also newly recorded in Fig.~\ref{fig:performance of case study}, and it can be seen from the figure that the introduced additional improved factors, \textit{i.e.}, the GAI models, inevitably increases training time. Moreover, we can observe that the training time of GDM-enhanced TD3 is significantly higher than that of the original TD3 algorithm. This is because the denoising time step in GDM leads to an increase in the training time of the neural network, while this also leads to a notable performance improvement. Similarly, although the training time of GDM-GAN-enhanced TD3 is slightly higher than that of the GDM-enhanced TD3, the former achieves better results, which is particularly beneficial for scenarios that require high decision accuracy.
\end{itemize}

\par Ultimately, all the GAI models are able to improve the performance of the original TD3 algorithm for dealing with the optimization problem in the case study. Moreover, more integration combination of GAI models in DRL can be explored in the future work.

\section{Future Directions}
\par In this section, we present some future directions for GAI-enhanced DRL.

\subsection{Efficiency Enhancement of GAI-enhanced DRL}
\par Although GAI models can improve the performance of DRL algorithms significantly, these models inevitably increases the computational complexity of the original DRL algorithms, which requires longer training time. Therefore, it is crucial to enhance the computational efficiency of GAI-enhanced DRL algorithms. One approach is to decompose the complex task into multiple simpler sub-tasks and distribute these sub-tasks across multiple machines or processors during the training process. This approach enhances efficiency by focusing on smaller and more manageable issues.


\subsection{GAI-assisted Reward Design}
\par As a key component of DRL, the quality of reward function design significantly impacts the performance of DRL. However, reward functions are usually manually set, and finding an appropriate reward function often requires multiple trials. In this context, the LLM integrated with retrieval-augmented generation (RAG) technique can offer a novel approach to improve the reward function design process. Specifically, the RAG model retrieves relevant information from existing reward function databases and research papers about DRL. Subsequently, the retrieved information is combined with the LLM to adjust and fine-tune the reward function for specific tasks or environments, thereby enhancing the accuracy of reward design.


\subsection{GAI-improved Transfer Learning}
\par If the application scenarios of DRL change, we usually have to retrain the DRL models from scratch, which results in a waste of computational resources. Notably, GAI shows great potential to solve this issue. Specifically, GAI can transfer the learned knowledge across different domains by identifying and exploiting common underlying structures that unify the distribution of newly generated data to a given distribution. Therefore, leveraging GAI to identify meta-knowledge that can be transferred across various domains helps bridge the gaps among different application scenarios of DRL~\cite{10515203}.

\subsection{Exploration of More GAI Models for DRL}
\par Different GAI models possess distinct characteristics and focuses. Therefore, it is essential to explore the integration of more GAI models into DRL algorithms to address various challenges faced by different DRL algorithms and improve the decision-making process. Based on this, refining the criteria for GAI model selection can ensure that the most suitable GAI model can be quickly chosen to address specific application scenarios and challenges within DRL. Moreover, it is necessary to develop appropriate evaluation metrics to assess the impact of GAI models on the performance of DRL algorithms.


\section{Conclusion}
\par In this paper, we have introduced how GAI can be applied to DRL to improve the performance. Specifically, we have first introduced the basic principles of several GAI models and DRL algorithms. Subsequently, we have discussed how GAI enhances DRL algorithms from both the data and policy perspectives. Then, we have proposed a framework to explore the improvement of DRL algorithm by four typical GAI models, namely GAN, GDM, VAE, and Transformer. Moreover, we have constructed a case study on near-field communication to validate the effectiveness of GAI-enhanced DRL. Experimental results have demonstrated that GDM shows the most significant enhancement for DRL. Finally, we have introduced four future directions to advance the wider application of GAI in DRL.

\bibliography{main}

\section*{Biographies}

\noindent 
\textsc{Geng Sun} (\text{sungeng@jlu.edu.cn}) received a B.S. degree in Communication Engineering from Dalian Polytechnic University, China, and the Ph.D. degree in Computer Science from Jilin University, China, in 2011 and 2018, respectively. He was a visiting researcher in the School of Electrical and Computer Engineering at Georgia Institute of Technology, USA. He is currently a Professor in the College of Computer Science and Technology at Jilin University, and his research interests include UAV Networks, collaborative beamforming, generative AI and optimizations.

\vspace{1em}

\noindent 
\textsc{Wenwen Xie} (\text{xieww22@mails.jlu.edu.cn}) received the B.S. degree in Computer Science and Technology from Hefei University of Technology, Hefei, China, in 2022. She is currently working toward the MS degree in Computer Science and Technology at Jilin University, Changchun, China. Her research interests include UAV communications, IRS beamforming and deep reinforcement learning.

\vspace{1em}

\noindent 
\textsc{Dusit Niyato} (\text{dniyato@ntu.edu.sg}) is currently a professor in the School of Computer Science and Engineering, Nanyang Technological University, Singapore. He received the B.Eng. degree from King Mongkut's Institute of Technology Ladkrabang (KMITL), Thailand, in 1999, and the Ph.D. in electrical and computer engineering from the University of Manitoba, Canada, in 2008. His research interests are in the areas of Internet of Things (IoT), machine learning, and incentive mechanism design.

\vspace{1em}

\noindent 
\textsc{Fang Mei} (\text{meifang@jlu.edu.cn}) received the M.Sc. and Ph.D. degrees in computer science from Jilin University, Changchun, China, in 2005 and 2010, respectively. She is currently an Associate Professor in the College of Computer Science and Technology at Jilin University. Her research interests include intelligent information processing, multi-access edge computing, vehicular networks, and antenna arrays.

\vspace{1em}

\noindent 
\textsc{Jiawen Kang} (\text{kavinkang@gdut.edu.cn}) received the Ph.D. degree from the Guangdong University of Technology, China, in 2018. He has been a postdoc at Nanyang Technological University, Singapore, from 2018 to 2021. He is currently a full professor at Guangdong University of Technology, China. His research interests focus on blockchain, security, and privacy protection.

\vspace{1em}

\noindent 
\textsc{Hongyang Du} (\text{duhy@hku.hk}) is an assistant professor at the Department of Electrical and Electronic Engineering, The University of Hong Kong. He received his Ph.D. degree from the College of Computing and Data Science, Energy Research Institute @ NTU, Nanyang Technological University, Singapore, in 2024. He serves as the Editor-in-Chief assistant of \textit{IEEE Communications Surveys \& Tutorials} (2022-2024), and the Guest Editor for \textit{IEEE Vehicular Technology Magazine}. His research interests include edge intelligence, generative AI, and network management.

\vspace{1em}

\noindent 
\textsc{Shiwen Mao} (\textit{smao@ieee.org}) received his Ph.D. in electrical  and computer engineering from Polytechnic University, Brooklyn, NY. He is a Professor and Earle C. Williams Eminent Scholar, and Director of the Wireless Engineering Research and Education Center at Auburn University. His research interests include wireless networks and multimedia communications.

\end{document}